\documentclass[pmlr]{jmlr}% new name PMLR (Proceedings of Machine Learning)

\RequirePackage{graphicx}
 \usepackage{booktabs}
 \usepackage{verbatim}

\usepackage{longtable}% for long tables
\usepackage{colortbl}
\usepackage{setspace}
\usepackage{enumitem}
\usepackage{makecell}

\makeatletter
\def\set@curr@file#1{\def\@curr@file{#1}} %temp workaround for 2019 latex release
\makeatother
\usepackage[load-configurations=version-1]{siunitx} % newer version

\newcommand{\repourl}{https://github.com/hyesunyun/llm-meta-analysis}

\theorembodyfont{\upshape}
\theoremheaderfont{\scshape}
\theorempostheader{:}
\theoremsep{\newline}

\jmlrvolume{252}
\jmlryear{2024}
\jmlrworkshop{Machine Learning for Healthcare}

\title[Automatically Extracting Numerical Results from RCTs with LLMs]{Automatically Extracting Numerical Results from Randomized Controlled Trials with Large Language Models}

\author{\Name{Hye Sun Yun}
       \Email{yun.hy@northeastern.edu}\\ 
       \addr
       Northeastern University\\
       Boston, MA, USA 
       \AND
       \Name{David Pogrebitskiy}
       \Email{pogrebitskiy.d@northeastern.edu}\\ 
       \addr
       Northeastern University\\
       Boston, MA, USA 
       \AND
       \Name{Iain J. Marshall}
       \Email{iain.marshall@kcl.ac.uk}\\ 
       \addr
       King's College London\\
       London, UK 
       \AND
       \Name{Byron C. Wallace}
       \Email{b.wallace@northeastern.edu}\\ 
       \addr
       Northeastern University\\
       Boston, MA, USA } 

\begin{document}

\maketitle

\begin{abstract}
  Meta-analyses statistically aggregate the findings of different randomized controlled trials (RCTs) to assess treatment effectiveness. Because this yields robust estimates of treatment effectiveness, results from meta-analyses are considered the strongest form of evidence. However, rigorous evidence syntheses are time-consuming and labor-intensive, requiring manual extraction of data from individual trials to be synthesized. Ideally, language technologies would permit fully automatic meta-analysis, on demand. This requires accurately extracting numerical results from individual trials, which has been beyond the capabilities of natural language processing (NLP) models to date. In this work, we evaluate whether modern large language models (LLMs) can reliably perform this task. We annotate (and release) a modest but granular evaluation dataset of clinical trial reports with numerical findings attached to interventions, comparators, and outcomes. Using this dataset, we evaluate the performance of seven LLMs applied zero-shot for the task of conditionally extracting numerical findings from trial reports. We find that massive LLMs that can accommodate lengthy inputs are tantalizingly close to realizing fully automatic meta-analysis, especially for dichotomous (binary) outcomes (e.g., mortality). However, LLMs---including ones trained on biomedical texts---perform poorly when the outcome measures are complex and tallying the results requires inference. This work charts a path toward fully automatic meta-analysis of RCTs via LLMs, while also highlighting the limitations of existing models for this aim.
\end{abstract}

\section{Introduction}

Quantitative measures of comparative treatment effectiveness are reported primarily in unstructured (natural language) published articles that describe the design, protocol, and results of randomized controlled trials (RCTs). 
Individual trial results are noisy and often biased, motivating the need for rigorous \emph{statistical meta-analysis} of all trials of a particular treatment to produce a robust estimate of efficacy \citep{lau1995cumulative,borenstein2021introduction}.\footnote{These are essentially weighted averages of comparative effect estimates reported in trials, where weights are inverse to reported variances.}
Estimates from meta-analyses of primary findings are considered one of the highest forms of evidence in medicine \citep{murad2016new}. 

However, conducting a meta-analysis requires manually extracting from individual articles the data elements necessary for synthesis, i.e., numbers required to calculate metrics of interest---for example, odds ratios or mean differences for dichotomous and continuous outcomes, respectively---and associated variances. 
This time-consuming but critical step still must be manually performed, which precludes a fully automated ``on-demand'' meta-analysis of all evidence relevant to a given clinical question.
Prior work \citep{yun2023appraising} has reported that domain experts who conduct systematic reviews and meta-analyses view data extraction as a key part of the process which may be amenable to automation via LLMs. 

In this work, we ask: \textbf{Are modern LLMs sufficiently capable of data extraction to permit accurate, fully automated meta-analysis?} 
To answer this question empirically, we annotate modest but granular validation and test sets of clinical trial reports with numerical findings attached to interventions, comparators, and outcomes (ICO triplets). 
We use these annotations to evaluate a wide range of LLMs---including accessible, smaller models and closed, massive models--- for their ability to infer structured numerical findings for specific comparisons of interest.
That is, we assess whether LLMs can reliably extract from trial reports the data necessary for statistical meta-analysis.

\begin{figure}%[b]
    \centering
    \includegraphics[scale=0.5]{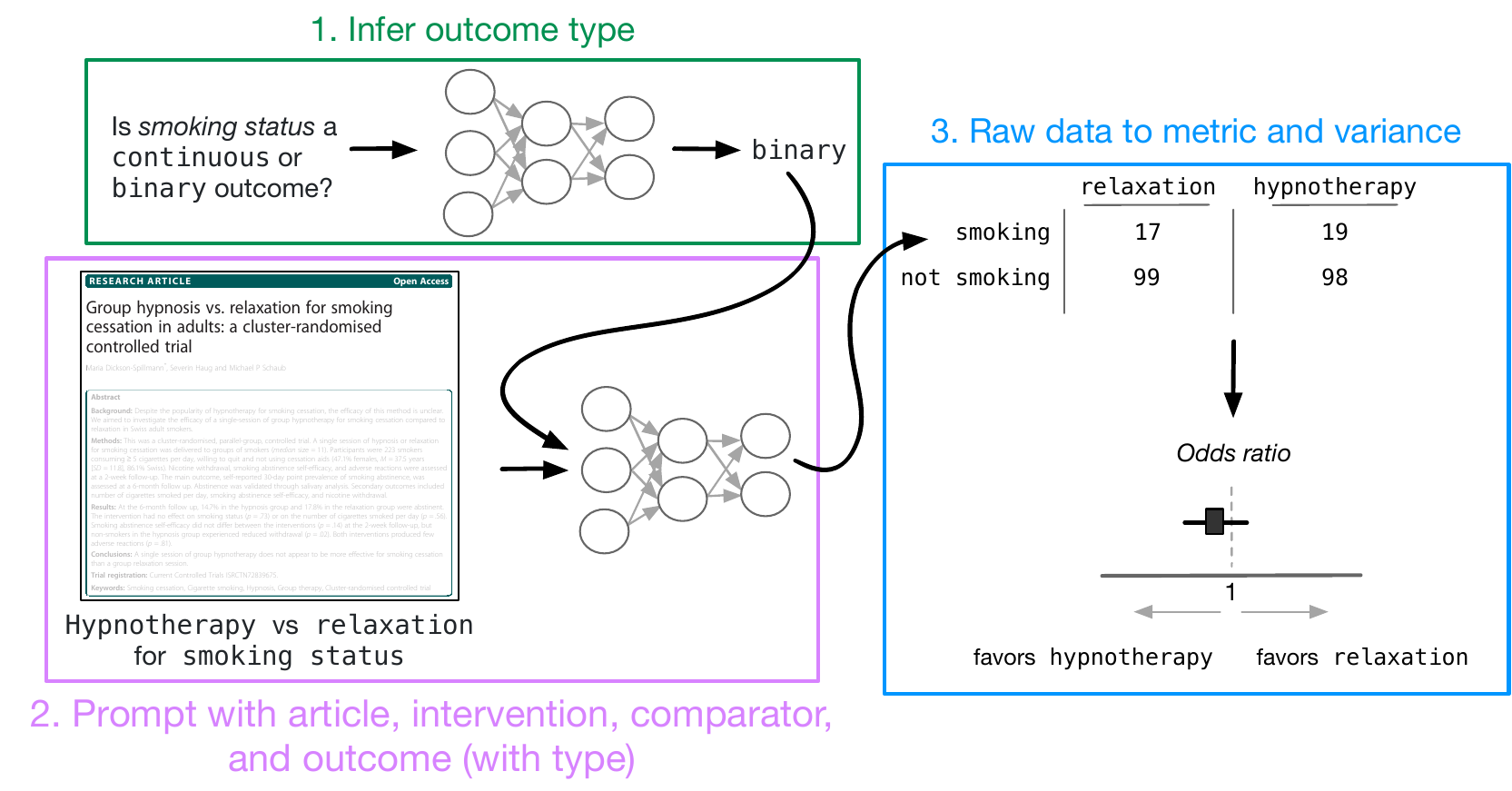}
    \caption{Overview of this work. (\textbf{\textcolor[HTML]{4f8c62}{1.}}) We use an LLM to infer a given outcome type based on its description, then (\textbf{\textcolor[HTML]{cf92e7}{2.}}) Prompt the LLM to extract raw data (e.g., number of participants who smoked in each treatment group). Finally, we (\textbf{\textcolor[HTML]{689ad8}{3.}}) transform this into a suitable metric (e.g., an odds ratio) and associated variance; this is the data necessary for statistical meta-analysis.}
    \label{fig:overview}
\end{figure}

We find that LLMs can permit accurate automated meta-analyses with some success. 
Unsurprisingly, massive LLMs with large input context windows such as GPT-4 outperform smaller, open-source models at extracting binary (dichotomous) outcomes. 
For extracting continuous outcomes, LLMs perform comparatively poorly, especially in settings where there are multiple similar outcome measures reported in a trial report. 
The takeaway from this work is that modern LLMs offer a promising path toward fully automatic meta-analysis, but further improvements are needed before this will be reliable. 
We hope the data we release with this effort will support future work in this direction. 

\subsection*{Generalizable Insights about Machine Learning in the Context of Healthcare}

LLMs have realized remarkable performance across a diverse range of NLP tasks in recent years, motivating several efforts to investigate the use of LLMs for healthcare specifically \citep{singhal2023large,lehman2023do,naik2023care,wadhwa2023jointly}. 
Here we provide empirical insights concerning the use of such models to fully automate meta-analysis of clinical trial results; the results also have implications for related numerical data extraction tasks in healthcare, and how well LLMs are likely to be able to perform them. 
More concretely, the main generalizable contributions this work offers are:
\begin{enumerate}[nosep]
    \item We release an evaluation dataset containing extensive annotations for the task of extracting numerical clinical findings for a given intervention, comparator, and outcome (ICO triplets) necessary for conducting meta-analyses.
    \item We report the quantitative and qualitative results evaluating modern LLMs on extracting numerical findings from RCTs using the annotated dataset. Furthermore, we provide an example of a fully automated meta-analysis to show the suitability of using LLMs for the end-to-end process.
\end{enumerate}

\noindent We hope that this effort highlights challenges that must be addressed to fully automate medical evidence synthesis, and ultimately brings us closer to that vision.

\section{Related Work}

Previous work has explored the use of NLP technologies to assist the process of meta-analyses and evidence synthesis. For example, there have been efforts to automate \emph{screening} to identify all studies relevant to a clinical question \citep{wallace2010active,kusa2024csmed}. 

Our focus, however, is on the \emph{data extraction}. 
In early work on automatic data extraction, \citet{kiritchenko2010exact} presented an automatic information extraction system called \texttt{ExaCT} that located and extracted key trial information such as sample size, drug dosage, and primary outcomes from full-text RCT articles. 
Elsewhere, \citet{summerscales2011automatic} proposed an automated way to create summaries from abstracts of RCTs by extracting numerical quantities of treatment groups and outcomes. 
Both of these efforts used statistical NLP models based on bag-of-words, as they pre-dated LLMs. 
Consequently, they achieved somewhat mediocre performance on this challenging task.

Do current-generation LLMs allow us to do much better? A few recent efforts have partially investigated this question. \citet{mutinda2022automatic} proposed a BERT \citep{devlin2018bert} based named entity recognition (NER) model to identify relevant trial information from research abstracts and parsed the numeric outcomes for statistical analysis. 
However, this work used an encoder-only model (BERT) which is considered small by modern standards and required explicit supervision for the task. 
Other recent work \citep{shamsabadi2024large} has shown that LLMs can be fine-tuned to perform information extraction tasks from scientific and biomedical literature in general, though these efforts were not focused on supporting meta-analysis. 

More recently, several works have explored extracting experimental findings of clinical trials using LLMs in zero-shot (a setup in which a model observes tasks or data that it has not explicitly seen during training) \citep{khraisha2023can,kartchner2023zero}. However, these works often evaluated a couple of massive, closed LLMs and did not focus on numerical findings. 
\citet{naik2023care} investigated numerical information extraction using several LLMs. However, theirs was a general extraction scheme intended to support a range of downstream functionalities; by contrast, we focus more narrowly on how well LLMs can extract the numerical data necessary for meta-analysis, conditioned on a specific intervention, comparator, and outcome of interest. 

The few focused efforts that have evaluated LLMs for data extraction from full-text articles for synthesis \citep{gartlehner2023data,sun2024good,reason2024artificial} have used very small convenience samples of articles and evaluated only closed models, providing a somewhat limited view of current model capabilities generally. In our work, we construct an evaluation dataset of several hundred samples with both abstracts and results sections from full-texts and evaluate numerical data extraction for meta-analyses across a diverse range of LLMs in zero-shot settings. 
In addition to massive, closed LLMs we evaluate smaller, accessible models to provide a more complete view of the capabilities of modern LLMs for numerical data extraction of RCTs.

\section{Extracting Numerical Results from Clinical Trial Reports}

We are interested in the task of inferring structured findings reported in clinical trial reports that correspond to a specific intervention, comparator, and outcome (ICO).
Past work \citep{lehman2019inferring,wadhwa2023jointly} has treated this as a \emph{classification} task, categorizing articles as reporting that the given intervention (e.g., aspirin) induced a \emph{significant increase}, \emph{significant decrease}, or \emph{no significant difference} with respect to the outcome of interest (e.g., duration of headache), relative to the specified comparator (e.g., placebo).

In this work, we consider a more challenging variant of this task in which the aim is to infer \emph{numerical} data associated with each ICO. 
Specifically, we investigate the ability of LLMs to extract the numbers required to derive point estimates (which capture the sign and magnitude of relative treatment effects) and corresponding variances. 
If we could reliably infer such results from individual articles describing trials, this would permit \emph{automatic statistical meta-analysis on demand}. 
Briefly, a standard fixed-effects meta-analysis---which assumes studies measure a common underlying effect \citep{hedges1998fixed}---aggregates point estimates of interest (e.g., odds ratios) $\hat{\theta}_i$ across $k$ studies (indexed by $i$) by taking a weighted sum, where weights $w_i$ are inverse to the variance associated with $\hat{\theta}_i$: 

\begin{equation}
    \hat{\theta}_{\text{FE}} = \frac{\sum_{i=1}^{k} w_i \hat{\theta}_i}{\sum_{i=1}^{k} w_i}
    \label{eq:FE}
\end{equation}

Meta-analysis of RCT evidence is critical because individual trials are inherently noisy and may reflect statistical biases.  
Aggregating evidence from independent trials in this way to estimate an overall treatment effect permits a robust effectiveness estimate \citep{egger1997meta}. 
For this reason, meta-analytic results provide reliable clinical evidence, and meta-analyses often inform healthcare policy and patient care guidelines.

A downside of meta-analyses is that they are laborious to produce and keep up to date with new findings, especially given the rate at which new evidence accumulates \citep{bastian2010seventy}.
This has motivated work on automating or semi-automating aspects of evidence synthesis, but reliably extracting the numerical values associated with ICO triplets has been a task too difficult for prior language technologies to reliably execute. 
Given the rapid progress in NLP---and specifically the capabilities of LLMs---we investigate, in this work, the degree to which modern LLMs can accurately extract numerical results from clinical trials sufficient for meta-analysis, fully automatically.

\section{Annotation}
\label{section:annotation}

To evaluate the feasibility of using LLMs to extract numerical data from RCTs for meta-analysis, we annotated a dataset of 699 abstracts and result sections from randomized controlled trials indexed in PubMed Central. 
Our dataset contains extensive annotations for the task of extracting numerical clinical findings for a given intervention, comparator, and outcome (ICO triplets) necessary for conducting meta-analyses (Equation \ref{eq:FE}).

Our dataset is derived from the \emph{Evidence Inference} corpus \citep{lehman2019inferring,deyoung2020evidence}.
This comprises articles describing RCTs.\footnote{Since it is constructed from articles in the open access subset, this dataset includes full-texts of papers in XML format.}
For all of these RCT reports, medical experts have assessed the directionality of the evidence reported concerning given ICO triplets.
Specifically, these annotations take the form of ternary classifications indicating whether the Intervention \emph{significantly increased}, \emph{significantly decreased}, or had \emph{no significant effect} on the Outcome, as compared to the Comparator. 
We enrich this dataset by attaching \emph{numerical} results to ICO triplets for a subset of the articles; these are at the level necessary to permit meta-analysis and so (much) more granular than the existing categorical annotations. 

\paragraph{Annotation Details} We randomly sampled 120 RCTs from the \emph{Evidence Inference} dataset \citep{deyoung2020evidence}. Full-text papers are quite lengthy. Therefore, as a practical matter, we focus on only abstracts and the results sections of the full-text papers, where numerical results are most often reported. 
We compress the XML-formatted sections of these papers to reduce input length.\footnote{We use \url{https://onlinexmltools.com/minify-xml}.}
We further pre-process the XML by removing all the attributes mostly about style rather than content. Finally, we convert each processed XML to markdown, further compressing the content while maintaining a vital organizational structure. We release the post-processed version of the data used in this work to ensure reproducibility.

During the annotation process, we made some changes to the Intervention, Comparator, and Outcomes fields for certain cases. For Outcome fields that included multiple parts, we divided the parts into separate individual Outcomes and associated them with the same Intervention and Comparator values. For example, if the original Outcome label was ``weight reduction at 12 months, BMI reduction at 12 months,'' then we created two separate Outcome values---one as ``weight reduction at 12 months'' and the other as ``BMI reduction at 12 months''---and associated these two new Outcomes with the same Intervention and Comparator. This results in having more ICO instances in the dataset than the original. In addition, we found cases when one Intervention field would have multiple interventions and the RCT report did not report numerical data for the combined, multiple interventions. In this case, we took a similar approach above and separated the parts into individual Interventions and associated them with the same Outcome and Comparator values. 

Finally for cases when the Outcome is associated with multiple numerical measurements in the RCT report, we transformed these Outcomes into multiple Outcomes with the specific measurement included in the value. For example, the Outcome of ``motion range'' had several measurements including shoulder internal rotation, shoulder external rotation, and shoulder extension. In this particular example, we created three separate records in the annotated dataset each with different Outcomes: ``motion range - shoulder internal rotation,'' ``motion range - shoulder external rotation,'' and ``motion range - shoulder extension.'' These derived Outcomes had the same Intervention and Comparator as the original Outcome.

We annotated each ICO from the dataset by following the annotation schema. The annotation schema included the following: 
\begin{enumerate}[nosep]
    \item \emph{Type of outcome}: binary or continuous
    \item \emph{Data extraction for binary outcomes}: intervention event number, intervention group size, comparator event number, and comparator group size (2x2 contingency table)
    \item \emph{Data extraction for continuous outcomes}: intervention mean, intervention standard deviation, intervention group size, comparator mean, comparator standard deviation, and comparator group size
\end{enumerate}

If a specific part of the numerical data is not found in the abstract or results section, the annotators left the field blank. We recorded metadata concerning whether the numerical information could be located within a figure or table and documented whether all of the desired numerical data was successfully retrieved.

To keep the task fairly straightforward, we decided to leave instances of RCTs that reported medians instead of means blank. If there are multiple time points for a given Outcome and the time point is not explicitly stated, we use the primary time point based on the RCT. If this is not provided, we use the last time point. Additionally, we only include the standard deviations if they were explicitly reported, or if the 95\% confidence intervals were provided, by back-calculating the standard deviations using the formula outlined in \citet{higgins2018standards}. 
We also report the token counts for each RCT abstract and results text (under OpenAI's tokenizer) to provide a sense of the input document length. 

We set aside 10 random RCTs as a development set for prompt engineering and used the remaining 120 RCTs as a test or evaluation set. We release this dataset via a GitHub repository.\footnote{\url{\repourl}}

\paragraph{Annotator backgrounds}
Two co-authors of the paper with backgrounds in computer science but familiarity with standard meta-analysis metrics annotated each record independently. 
The annotators met regularly to agree on the final annotations. 
We also solicited feedback from a clinical researcher with expertise in evidence synthesis to discuss edge cases and ensure annotation quality. 

\paragraph{Dataset statistics}

\begin{table}[t]
\small
\centering
\caption{Statistics for the annotated dataset.}
\label{tab:dataset}
\begin{tabular}{lccc}
\toprule
\textbf{Metric}          & \textbf{Dev} & \textbf{Test} & \textbf{Total} \\ \midrule
\rowcolor[HTML]{EFEFEF} 
\# PMC Articles          & 10           & 110           & 120            \\
\# Prompts (ICOs)        & 43           & 656           & 699            \\
\rowcolor[HTML]{EFEFEF} 
\# Binary Outcomes       & 11           & 172           & 183            \\
\# Continuous Outcomes   & 32           & 484           & 516            \\
\rowcolor[HTML]{EFEFEF} 
\% With Enough Data for Point Estimates   & 62.79          & 58.84           & 59.08           \\
Mean Articles Tokens & 3331        & 3603         & 3581          \\ \bottomrule
\end{tabular}
\end{table}

\autoref{tab:dataset} gives an overview of statistics for our final dataset. Our dataset includes 120 RCTs (abstracts and results sections) and 699 total records. The development set has 10 RCTs with 43 records and the test set has 110 RCTs with 656 records. A total of 183 outcomes are categorized as binary while 516 as continuous. We found that a total of 413 records (59.08\%) have sufficient data to calculate the point estimates of the clinical trials. The numerical data extracted during annotations were mostly found in the tables of the RCTs as 471 instances fell under this case. The average token number for each RCT was 3,581.

\section{Can LLMs Accurately Extract Numerical Results from Trial Reports?} 

We evaluate whether modern LLMs are capable of extracting the numerical data necessary for meta-analysis from RCT reports. We assume a \emph{zero-shot} setting here, i.e., we instruct models to extract the data elements of interest, without additional supervision. 
Few-shot learning may realize better performance for this task. 
However, RCT reports are often quite lengthy, even when considering only abstracts and results sections.
This makes it practically difficult (and expensive) to include full examples in context. Therefore, we evaluate only the zero-shot case in this work.

Ideally, one might ask directly for the \emph{point estimate} of interest, for example, the odds ratio with respect to a (dichotomous) outcome of interest between the treatment and control groups. 
However, deriving this often requires intermediate steps to transform raw data (e.g., a two-by-two table tallying the number of participants in each group who did and did not experience a specific outcome) into point estimates (e.g., odds ratios) and standard errors.

Given that LLMs are middling at math \citep{hong2024caught, satpute2024llms, urrutia2024s}---and the fact that statistical meta-analysis packages like \texttt{metafor} \citep{metafor} will readily compute such quantities from raw data---we instead adopt a stepwise approach intended to extract raw data.
Specifically, we first prompt the model to infer whether a given outcome is binary or continuous based on a (natural language) description of the outcome, e.g, \emph{elevation of glucose after 1 hour}; these descriptions are part of the annotations provided in the evidence inference dataset \citep{deyoung2020evidence}.
Note that this assessment is made independently of the input article, and in some cases, the model may respond that it is unable to make this inference. 
The output from this step informs whether we prompt the model to extract raw data as would be reported in the dichotomous case (i.e., a two-by-two table tallying numbers of events in the treatment and control groups, respectively) and continuous instance (measurements and variances associated with the outcome of interest in both groups).

Due to the limited sequence lengths of many smaller, open-source models, it was necessary to divide texts into segments sufficiently small to conform to a maximum length. 
This \emph{chunking} process involves breaking down texts into smaller sections. 
Our approach includes pre-processing the text, converting number words to digits, and removing sentences without numerical values. 
We then iteratively concatenate segments until we have a set of chunks that are together at the token limit.

We pass the raw data extracted from this step through specialized statistical software (\texttt{statsmodels}; \citealt{seabold2010statsmodels} or \texttt{metafor}; \citealt{metafor}) to derive point estimates and standard errors, i.e., the inputs necessary for meta-analysis. 
This sequential approach provides a form of transparency: Users can inspect the ``raw'' outputs from LLMs to verify their correctness. 
We release the code and prompts used for the evaluation: \url{\repourl}.

\subsection{Large Language Models (LLMs)}

\begin{table}[t]
\centering
\caption{Full list of instruction-tuned LLMs used for the evaluation experiments.}
\label{tab:models}
\resizebox{\textwidth}{!}{%
\begin{tabular}{lcccl}
\toprule
\textbf{Model}               & \textbf{Type}        & \textbf{Parameters} & \textbf{Sequence Length} & \textbf{Training Data}                                                                                            \\ \midrule
\rowcolor[HTML]{EFEFEF} 
\textbf{GPT-4 0125}          & General     & 1.7T       & 128000          & Unknown                                                                                                  \\
\textbf{GPT-3.5 Turbo 0125}  & General     & 175B       & \phantom{0}16385           & Unknown                                                                                                  \\
\rowcolor[HTML]{EFEFEF} 
\textbf{Alpaca} & General     & 13B         & \phantom{0}\phantom{0}4096           & \begin{tabular}[c]{@{}l@{}}instruction-following demonstrations\\generated from OpenAI’s \texttt{text-davinci-003}\end{tabular}                                                                                          \\
\textbf{Mistral Instruct v2} & General     & 7B         & \phantom{0}32768           & Unknown                                                                                                  \\
\rowcolor[HTML]{EFEFEF} 
\textbf{Gemma Instruct}      & General     & 7B         & \phantom{0}\phantom{0}8192            & \begin{tabular}[c]{@{}l@{}}Web Documents, Code, \\ Mathematical text, Instructions\end{tabular}          \\
\textbf{OLMo Instruct}       & Science     & 7B         & \phantom{0}\phantom{0}2048            & \begin{tabular}[c]{@{}l@{}}Tulu 2 SFT Mix, \\ Ultrafeedback Cleaned\end{tabular}                         \\
\rowcolor[HTML]{EFEFEF} 
\textbf{PMC LLaMA}           & Biomedicine & 13B        & \phantom{0}\phantom{0}2048            & \begin{tabular}[c]{@{}l@{}}Semantic Scholar Open Research Corpus, \\ PMC LLaMA instructions\end{tabular} \\
\textbf{BioMistral}          & Biomedicine & 7B         & \phantom{0}\phantom{0}2048            & PubMed Central Open Access Subset                                                                        \\ \bottomrule
\end{tabular}%
}
\end{table}

Recent work has demonstrated that LLMs are strong entity and relation extractors. 
\citet{wadhwa2023revisiting} showed that LLMs can achieve comparable results to fully supervised models for relation extraction, zero-shot. 
Related efforts have demonstrated the capabilities of zero-shot LLMs in healthcare specifically \citep{wei2021finetuned, agrawal2022large, singhal2023large, sivarajkumar2024empirical}.
In this work, we evaluate the following LLMs with respect to their ability to extract the data from RCT reports necessary for meta-analysis:
\begin{itemize}[leftmargin=*,topsep=0pt]
\setlength\itemsep{-0.5em}
    \item \textbf{GPT-4} \citep{achiam2023gpt}: The latest version of the GPT-family of models from OpenAI. This is a large multimodal (text and image) model. We use \verb|gpt-4-0125| which has a maximum context length of 128k. Context length is the amount of text that an LLM can process and retain in memory at any given time.
    \item \textbf{GPT-3.5}\footnote{\url{https://platform.openai.com/docs/models/gpt-3-5-turbo}}: Also from OpenAI, the model behind ChatGPT. 
    We use  \verb|gpt-3.5-turbo-0125| which can handle contexts of up to 16,385 tokens.
    \item \textbf{Alpaca 13B} \citep{alpaca}: An instruction fine-tuned version of the LLaMA-13B \citep{touvron2023llama} model with a max context length of 4,096 tokens.
    \item \textbf{Mistral 7B Instruct v2}\footnote{\url{https://huggingface.co/mistralai/Mistral-7B-Instruct-v0.2}}: An instruction fine-tuned version of the Mistral-7B-v0.1 \citep{jiang2023mistral} model. Allows for a context length of 32,768 tokens.
    \item \textbf{Gemma 7B Instruct} \citep{team2024gemma}: An open-weights LLM from Google DeepMind. Max context length: 8,192 tokens.
    \item \textbf{OLMo 7B Instruct} \citep{groeneveld2024olmo}: Open Language Model (OLMo) from AI2. Features an input sequence length upper bound of 2,048 tokens.
    \item \textbf{PMC LLaMa} \citep{wu2023pmc}: A 13B parameter model which was initialized to LLaMA-13B \citep{touvron2023llama} and then further pretrained over medical corpora. The model was then instruction-tuned. It offers a max input and output sequence length of 2,048.
    \item \textbf{BioMistral} \citep{labrak2024biomistral}: A Mistral-based model which was further pre-trained over data from PubMed Central Open Access. It has an input upper bound of 2,048 tokens (even though the base model has a significantly greater sequence length; presumably this was done due to resource constraints). 
\end{itemize}

\subsection{Metrics}

To measure the performance of LLMs in terms of their ability to extract numerical data from trial reports, we first consider their accuracy with respect to categorizing outcome types (as binary or continuous), and we then evaluate data extraction accuracy both in terms of exact and partial matches with reference (manually extracted) data.

We consider data extracted from a trial report partially correct where a subset of the extracted numerical data matches the reference. 
For example, for a binary outcome with 4 numerical values (two-by-two table entries), correctly extracting 1, 2, or 3 of these may be viewed as a partial match.
We calculate and report partial accuracy for all possible partial matches.

Articles sometimes report results ambiguously, or not at all. In these cases, we wish the model to abstain from providing data. We also report the number of times the model mistakenly outputs ``unknown'' for a specific label. 
This measure is useful for understanding how conservative the model is, as it captures instances where humans do not find the results ambiguous but the models do.

In addition to considering whether extracted data matches reference data (one binary indicator per ICO), we consider the magnitude of difference between extracted values and references (in standardized units). 
This is informative as it indicates whether errors are likely to have an important impact on downstream meta-analysis (large magnitude differences would affect the overall treatment efficacy estimate $\hat{\theta}_{\text{FE}}$ more). 
We calculate mean standardized errors (average of all the differences between the efficacy estimates from numerical data of LLMs and reference) and their variance (standard error and 95\% confidence interval).

\section{Results}
\label{sec:results}

We evaluated the performance of LLMs on three different tasks that are aimed to assist in automating meta-analyses in zero-shot. The three tasks are inferring outcome type, extracting numerical results for binary outcomes, and extracting numerical results for continuous outcomes. 
These tasks were performed independently of each other. LLMs were tasked to output the answers in a structured format (categorical answer for inferring outcome type and YAML format for data extraction). 
Lastly, we show the promise of using LLMs end-to-end for automating meta-analyses with a case study where we reproduce a meta-analysis with fully automated data.

\subsection{Extracting Data from Trial Reports}

\begin{table}[t]
\centering
\caption{LLM performances for inferring outcome type (binary or continuous). \textbf{\# Unknowns} refers to instances when the model outputs the unknown token ``x''; these are effectively incorrect (here the type should be inferable).}
\label{tab:outcome_type}
\resizebox{\textwidth}{!}{%
\begin{tabular}{ccccccccc}
\toprule
\rowcolor[HTML]{FFFFFF} 
\textbf{} &
  \textbf{GPT-4} &
  \textbf{GPT-3.5} &
  \textbf{Alpaca} &
  \textbf{Mistral} &
  \textbf{Gemma} &
  \textbf{OLMo} &
  \textbf{\thead{PMC\\LLaMA}} &
  \textbf{BioMistral} \\ \midrule
\rowcolor[HTML]{EFEFEF} 
\multicolumn{1}{c|}{\cellcolor[HTML]{FFFFFF}\textbf{Accuracy}} & 0.713          & 0.607 &  \textbf{0.739} & 0.201 & 0.665      & 0.290 & 0.732 & 0.133\\ \midrule
\rowcolor[HTML]{FFFFFF} 
\multicolumn{1}{c|}{\cellcolor[HTML]{FFFFFF}\textbf{F1 - Binary}} & \textbf{0.735} & 0.680 & 0.000 & 0.576 & 0.590      & 0.424 & 0.124          & 0.275 \\
\rowcolor[HTML]{EFEFEF} 
\multicolumn{1}{c|}{\cellcolor[HTML]{FFFFFF}\textbf{F1 - Continuous}} &
  0.836 &
  0.690 &
  \textbf{0.851} &
  0.183 &
  0.716 &
  0.079 &
  0.848 &
  0.135 \\ \midrule
\rowcolor[HTML]{FFFFFF} 
\multicolumn{1}{c|}{\cellcolor[HTML]{FFFFFF}\textbf{\# Unknowns}} & 155 & 152   & 1 & 489   & \textbf{0} & 5     & 15             & 409   \\ \bottomrule
\end{tabular}%
}
\end{table}

\paragraph{Inferring outcome types} Results for inferring outcome types are reported in \autoref{tab:outcome_type}. Alpaca-13B realizes the highest exact match accuracy of 0.739, followed by PMC LLaMA with 0.732. Although Alpaca-13B and PMC LLaMA have high accuracy, their F1 scores for the binary label are poor because the model mostly predicts the outcome type to be ``continuous'', which is the majority class (74.78\% of the test set).
GPT-4 and GPT-3.5 achieve high accuracy and F1 scores, besting the open-source models. 
The open, smaller models we evaluated appear less stable, i.e., the F1 score varies considerably.
Among these models, Gemma offers the best performance for exact match accuracy, F1 scores, and the number of instances where the model designates the outcome type as ``unknown''.

\begin{table}[t]
\centering
\caption{Performance in terms of extracting numerical values for binary outcomes. \textbf{IE}: intervention events; \textbf{IGS}: intervention group size; \textbf{CE}: comparator events, and; \textbf{CGS} comparator group size. Partial match numbers reflect leniency in the number of matches required for an instance to count as (partially) ``correct''. The MSE is the mean standardized error of the log odds ratios calculated with data extracted from the model. The number of unknowns refers to the number of times the model produced the unknown token ``x'' when the reference was not unknown; in this case, the reference data contained unknowns 72 times. The percentage of complete data shows how many model outputs had enough data to calculate the point estimate (and variance), compared to the reference data.}
\label{tab:binary_outcomes}
\resizebox{\textwidth}{!}{%
\begin{tabular}{cccccccccc}
\toprule
\rowcolor[HTML]{FFFFFF} 
\textbf{} &
   &
  \textbf{GPT-4} &
  \textbf{GPT-3.5} &
  \textbf{Alpaca} &
  \textbf{Mistral} &
  \textbf{Gemma} &
  \textbf{OLMo} &
  \textbf{\thead{PMC\\LLaMA}} &
  \textbf{BioMistral} \\ \midrule
\rowcolor[HTML]{EFEFEF} 
\multicolumn{1}{c|}{\cellcolor[HTML]{FFFFFF}\textbf{}} &
  \multicolumn{1}{c|}{\cellcolor[HTML]{FFFFFF}\emph{\textbf{Total}}} &
  \textbf{0.655} &
  0.298 &
  0.035 &
  0.164 &
  0.135 &
  0.012 &
  0.035 &
  0.035 \\
\rowcolor[HTML]{FFFFFF} 
\multicolumn{1}{c|}{\textbf{}} &
  \multicolumn{1}{c|}{\emph{\textbf{IE}}} &
  \textbf{0.749} &
  0.462 &
  0.129 &
  0.345 &
  0.275 &
  0.076 &
  0.146 &
  0.158 \\
\rowcolor[HTML]{EFEFEF} 
\multicolumn{1}{c|}{\cellcolor[HTML]{FFFFFF}\textbf{Exact Match}} &
  \multicolumn{1}{c|}{\cellcolor[HTML]{FFFFFF}\emph{\textbf{IGS}}} &
  \textbf{0.842} &
  0.655 &
  0.094 &
  0.515 &
  0.509 &
  0.170 &
  0.088 &
  0.053 \\
\rowcolor[HTML]{FFFFFF} 
\multicolumn{1}{c|}{\textbf{}} &
  \multicolumn{1}{c|}{\emph{\textbf{CE}}} &
  \textbf{0.737} &
  0.392 &
  0.129 &
  0.333 &
  0.275 &
  0.123 &
  0.158 &
  0.158 \\
\rowcolor[HTML]{EFEFEF} 
\multicolumn{1}{c|}{\cellcolor[HTML]{FFFFFF}\textbf{}} &
  \multicolumn{1}{c|}{\cellcolor[HTML]{FFFFFF}\emph{\textbf{CGS}}} &
  \textbf{0.830} &
  0.649 &
  0.094 &
  0.567 &
  0.556 &
  0.140 &
  0.058 &
  0.053 \\ \midrule
\rowcolor[HTML]{FFFFFF} 
\multicolumn{1}{c|}{\textbf{}} &
  \multicolumn{1}{c|}{\emph{\textbf{3}}} &
  \textbf{0.690} &
  0.415 &
  0.035 &
  0.251 &
  0.216 &
  0.035 &
  0.035 &
  0.035 \\
\rowcolor[HTML]{EFEFEF} 
\multicolumn{1}{c|}{\cellcolor[HTML]{FFFFFF}\textbf{Partial Match}} &
  \multicolumn{1}{c|}{\cellcolor[HTML]{FFFFFF}\emph{\textbf{2}}} &
  \textbf{0.901} &
  0.696 &
  0.181 &
  0.637 &
  0.585 &
  0.187 &
  0.164 &
  0.175 \\
\rowcolor[HTML]{FFFFFF} 
\multicolumn{1}{c|}{\textbf{}} &
  \multicolumn{1}{c|}{\emph{\textbf{1}}} &
  \textbf{0.912} &
  0.749 &
  0.193 &
  0.708 &
  0.678 &
  0.275 &
  0.216 &
  0.175 \\ \midrule
\rowcolor[HTML]{EFEFEF} 
\multicolumn{2}{c|}{\cellcolor[HTML]{FFFFFF}\textbf{MSE}} &
  \textbf{0.101} &
  0.441 &
  0.485 &
  0.657 &
  0.913 &
  1.253 &
  1.523 &
  - \\ \midrule
\rowcolor[HTML]{FFFFFF} 
\multicolumn{2}{c|}{\textbf{\# Unknowns}} &
  41 &
  145 &
  490 &
  \textbf{28} &
  90 &
  319 &
  524 &
  612 \\ \midrule
\rowcolor[HTML]{EFEFEF} 
\multicolumn{2}{c|}{\cellcolor[HTML]{FFFFFF}\textbf{\% Complete}} &
  \textbf{87.94} &
  61.70 &
  9.22 &
  87.23 &
  58.87 &
  24.11 &
  7.09 &
  0.00 \\ \bottomrule
\end{tabular}%
}
\end{table}

\paragraph{Binary outcomes} Performance metrics for binary outcome data extraction are available in \autoref{tab:binary_outcomes}. Based on the accuracy metrics and number of complete data, GPT-4 significantly outperforms all the other models. GPT-4 has the highest overall exact match accuracy of 0.655, followed by GPT-3.5 with 0.298. Similar to the task of inferring outcome types, the open, smaller models vary substantially in terms of their match accuracies. Especially, the models trained on biomedical text have one of the worst performances with both PMC LLaMA and BioMistral achieving a mere 0.035 in total exact accuracy and producing the most number of ``unknowns'' leading to the lowest percentage of complete data. 

The mean standardized errors of the log odds ratio calculated from the extracted data are lowest for the best performing models and highest for the worst performing models. GPT-4's MSE is 0.101 with a standard error of 0.043 (95\% CI 0.017 to 0.186) while BioMistral did not extract any computable data comparable with our reference data.

\begin{table}[t]
\centering
\caption{Performance of extracting numerical values for continuous outcomes. \textbf{IM}: intervention mean, \textbf{ISD}: intervention standard deviation, \textbf{IGS}: intervention group size, \textbf{CM}: comparator mean, \textbf{CSD}: comparator standard deviation, and \textbf{CGS}: comparator group size. Partial match numbers refer to how many parts need to match the reference for the instance to count as correct. The MSE is the mean standardized error of the standardized mean differences calculated with data extracted from the model. The number of unknowns refers to the number of times the model produced the unknown token ``x'' when the reference was not unknown; in this case, the reference data contained unknowns 925 times. The percentage of complete data shows how many of the model outputs had enough data to calculate the point estimate (and
variance), compared to the reference data.}
\label{tab:continuous_outcomes}
\resizebox{\textwidth}{!}{%
\begin{tabular}{cccccccccc}
\toprule
\rowcolor[HTML]{FFFFFF} 
\textbf{} &
  \emph{\textbf{}} &
  \textbf{GPT-4} &
  \textbf{GPT-3.5} &
  \textbf{Alpaca} &
  \textbf{Mistral} &
  \textbf{Gemma} &
  \textbf{OLMo} &
  \textbf{\thead{PMC\\LLaMA}} &
  \textbf{BioMistral} \\ \midrule
\rowcolor[HTML]{EFEFEF} 
\multicolumn{1}{c|}{\cellcolor[HTML]{FFFFFF}\textbf{}} &
  \multicolumn{1}{c|}{\cellcolor[HTML]{FFFFFF}\emph{\textbf{Total}}} &
  \textbf{0.487} &
  0.280 &
  0.039 &
  0.095 &
  0.087 &
  0.035 &
  0.039 &
  0.041 \\
\rowcolor[HTML]{FFFFFF} 
\multicolumn{1}{c|}{\textbf{}} &
  \multicolumn{1}{c|}{\emph{\textbf{IM}}} &
  \textbf{0.720} &
  0.538 &
  0.309 &
  0.348 &
  0.328 &
  0.221 &
  0.369 &
  0.390 \\
\rowcolor[HTML]{EFEFEF} 
\multicolumn{1}{c|}{\cellcolor[HTML]{FFFFFF}\textbf{}} &
  \multicolumn{1}{c|}{\cellcolor[HTML]{FFFFFF}\emph{\textbf{ISD}}} &
  \textbf{0.751} &
  0.606 &
  0.334 &
  0.375 &
  0.412 &
  0.311 &
  0.447 &
  0.470 \\
\rowcolor[HTML]{FFFFFF} 
\multicolumn{1}{c|}{\textbf{Exact Match}} &
  \multicolumn{1}{c|}{\emph{\textbf{IGS}}} &
  \textbf{0.734} &
  0.641 &
  0.216 &
  0.507 &
  0.534 &
  0.190 &
  0.107 &
  0.087 \\
\rowcolor[HTML]{EFEFEF} 
\multicolumn{1}{c|}{\cellcolor[HTML]{FFFFFF}\textbf{}} &
  \multicolumn{1}{c|}{\cellcolor[HTML]{FFFFFF}\emph{\textbf{CM}}} &
  \textbf{0.720} &
  0.526 &
  0.330 &
  0.361 &
  0.324 &
  0.227 &
  0.390 &
  0.402 \\
\rowcolor[HTML]{FFFFFF} 
\multicolumn{1}{c|}{\textbf{}} &
  \multicolumn{1}{c|}{\emph{\textbf{CSD}}} &
  \textbf{0.738} &
  0.584 &
  0.338 &
  0.390 &
  0.404 &
  0.282 &
  0.456 &
  0.472 \\
\rowcolor[HTML]{EFEFEF} 
\multicolumn{1}{c|}{\cellcolor[HTML]{FFFFFF}\textbf{}} &
  \multicolumn{1}{c|}{\cellcolor[HTML]{FFFFFF}\emph{\textbf{CGS}}} &
  \textbf{0.691} &
  0.608 &
  0.181 &
  0.427 &
  0.447 &
  0.184 &
  0.109 &
  0.087 \\ \midrule
\rowcolor[HTML]{FFFFFF} 
\multicolumn{1}{c|}{\textbf{}} &
  \multicolumn{1}{c|}{\emph{\textbf{5}}} &
  \textbf{0.542} &
  0.336 &
  0.045 &
  0.115 &
  0.103 &
  0.060 &
  0.058 &
  0.054 \\
\rowcolor[HTML]{EFEFEF} 
\multicolumn{1}{c|}{\cellcolor[HTML]{FFFFFF}} &
  \multicolumn{1}{c|}{\cellcolor[HTML]{FFFFFF}\emph{\textbf{4}}} &
  \textbf{0.724} &
  0.555 &
  0.293 &
  0.293 &
  0.342 &
  0.173 &
  0.375 &
  0.402 \\
\rowcolor[HTML]{FFFFFF} 
\multicolumn{1}{c|}{\textbf{Partial Match}} &
  \multicolumn{1}{c|}{\emph{\textbf{3}}} &
  \textbf{0.765} &
  0.645 &
  0.311 &
  0.421 &
  0.408 &
  0.231 &
  0.392 &
  0.408 \\
\rowcolor[HTML]{EFEFEF} 
\multicolumn{1}{c|}{\cellcolor[HTML]{FFFFFF}} &
  \multicolumn{1}{c|}{\cellcolor[HTML]{FFFFFF}\emph{\textbf{2}}} &
  \textbf{0.913} &
  0.814 &
  0.470 &
  0.691 &
  0.699 &
  0.408 &
  0.497 &
  0.501 \\
\rowcolor[HTML]{FFFFFF} 
\multicolumn{1}{c|}{} &
  \multicolumn{1}{c|}{\emph{\textbf{1}}} &
  \textbf{0.922} &
  0.872 &
  0.551 &
  0.794 &
  0.810 &
  0.507 &
  0.518 &
  0.501 \\ \midrule
\rowcolor[HTML]{EFEFEF} 
\multicolumn{2}{c|}{\cellcolor[HTML]{FFFFFF}\textbf{MSE}} &
  \textbf{0.290} &
  0.951 &
  6.257 &
  1.138 &
  3.466 &
  1.738 &
  - &
  - \\ \midrule
\rowcolor[HTML]{FFFFFF} 
\multicolumn{2}{c|}{\textbf{\# Unknowns}} &
  \textbf{422} &
  437 &
  1169 &
  483 &
  775 &
  1213 &
  1778 &
  1985 \\ \midrule
\rowcolor[HTML]{EFEFEF} 
\multicolumn{2}{c|}{\cellcolor[HTML]{FFFFFF}\textbf{\% Complete}} &
  \textbf{63.64} &
  62.40 &
  31.82 &
  62.81 &
  40.08 &
  11.98 &
  4.96 &
  0.00 \\ \bottomrule
\end{tabular}%
}
\end{table}

\paragraph{Continuous outcomes} \autoref{tab:continuous_outcomes} presents the performance metrics for continuous outcome data extraction. Due to this task involving more values to extract, LLM performance was quite low with even GPT-4 achieving exact match accuracy of 0.487. The second best model is GPT-3.5 with a total exact match accuracy dropping to 0.280. Similarly to binary outcome data extraction, the smaller, open-source models performed poorly. 

The mean standardized errors show the comparison between the effect estimates calculated from model outputs and the reference data. GPT-4 has the lowest value of 0.290 (SE: 0.112; 95\% CI: 0.071 to 0.510), followed by GPT-3.5 with 0.951 (SE: 0.420, 95\% CI: 0.127 to 1.775). LLMs further trained on biomedical texts consistently performed the worst with BioMistral producing no complete data to calculate the standardized mean difference.

\paragraph{Error Analysis} To characterize the types of errors that LLMs make when extracting numerical values from trial reports, we qualitatively evaluated mistakes observed on validation data. 
For this analysis, we focused on outputs from GPT-4 and Mistral as representative models; these were the best-performing closed and open LLMs evaluated, respectively. 

For outcome type inference, GPT-4 produced an ``unknown'' a total of 155 times, inferred an outcome to be continuous when it was in fact binary 25 times, and predicted a continuous outcome as binary 8 times. 
Mistral 7B Instruct had 480, 34, and 1 errors, respectively. 
In addition, Mistral produced 9 badly formatted YAML outputs.

For binary outcome extraction, GPT-4 errors broke down as follows: Producing incorrect numbers (34); Producing ``unknown'' for inferable values (20); and Outputting actual numbers when the reference is ``unknown'' (12). 
Mistral 7B Instruct yielded invalid output formats 3 times. In addition, Mistral produced incorrect numerical values (123); ``unknown'' values when they should not have (7); and confabulated numerical values for ``unknown'' reference values (25). 

For continuous outcome extraction, the majority of errors involved producing the wrong numerical value (GPT-4: 113; Mistral: 226); producing ``unknown'' values inappropriately (GPT-4: 142; Mistral: 171); and generating numerical values for ``unknown'' reference values (GPT-4: 38; Mistral: 171). Mistral also produced 5 poorly formatted outputs in this case. \nameref{app:error_plots} visualizes the types and number of errors from GPT-4 and Mistral.

We sampled 136 errors from GPT-4 and 148 errors from Mistral. We sampled up to 20 errors from each general error category mentioned above. Then, we conducted a qualitative error analysis to better characterize the mistakes made by these LLMs. We inductively annotated possible sources and reasons for each sampled error and aggregated commonly found reasons. Selected samples of errors are available in \nameref{app:samples}.

In addition to LLMs producing the wrong labels for clearly defined outcomes, our analysis revealed that errors in inferring outcome type are often attributable to ambiguous phrasing of the outcome values. 
Since the additional context of the trial reports is not provided as input, some outcome values can be either reported as a binary outcome or continuous. Examples of this include words such as ``rate'' and ``level'' even in cases where they are treated as dichotomous, and ``failure'' and ``discontinuation'' even when they are continuous.
This suggests that providing additional context from the article when making this classification may improve performance. 

For outcome extraction tasks, we found that smaller models such as Mistral sometimes produced formats that do not conform to the expected outputs whereas GPT-4 did not have this problem. 
These bad formats were often the contributing factor for having more ``unknowns'' since badly formatted outputs were treated as ``unknown'' numerical data.

We observed that incorrect outputs were often due to ``hallucinations'' (outputting numerical data not found anywhere in the trial report) or extracting values from the wrong intervention, comparator, or outcome measures from the wrong timepoint. Also, models often produced ``unknown'' as an answer if the target numerical value was a negative value. One possible explanation can be that these negative numbers are often reported in RCTs with two hyphens (``\verb|--|'') rather than a single one preceding a number which can lead to the model ignoring them as proper numerical values.
When RCTs report pre- and post-treatment measures, LLMs sometimes extract the values only from the pre-treatment and ignore the post-treatment numbers (whereas we need both for meta-analysis; typically one is interested in the comparative differences). 

One common apparent source of error occurs when the LLMs need to perform simple math such as division or subtraction to get the correct total group sizes. 
For example, an abstract might report that 40 participants were randomly assigned to two study groups. In this case, most humans would assume (absent additional information) that each group was assigned 20 participants. However, LLMs sometimes fail to make such inferences and output totally irrelevant numbers or ``unknown.'' 
Another common group size mistake appears to be due to the trials containing several different group sizes which correspond to different ICO triplets. Some outcome measures are reported based on the sample that successfully finished the study, for example. Finally, while apparently rare, LLMs sometimes confuse medians and means, or standard errors and standard deviations. 

\subsection{Towards Fully Automatic Meta-analysis: An Example} 

To demonstrate the feasibility of conducting a standard (fixed effects) meta-analysis with LLMs, we provide a case study based on an existing meta-analysis published in the Cochrane Library on remdesivir for treatment of COVID-19 \citep{grundeis2023remdesivir}. 
For this exercise, we focus on one outcome: \emph{all-cause mortality at up to day 28}. 
This outcome is a dichotomous (binary) outcome, and the comparison is between remdesivir as the intervention and a comparator of standard care. 
The original meta-analysis includes four trial reports and reports all important numerical values extracted from each trial.

We followed the same pre-processing approach for preparing the trial reports after downloading their full text in XML via \texttt{Bio.Entrez}\footnote{\url{https://biopython.org/docs/1.75/api/Bio.Entrez.html}} Python package \citep{cock2009biopython}. 
We prompted both GPT-4 and Mistral Instruct 7B to extract the numerical values.
We start by inferring the outcome type, and then we use the outputs from this step to prompt the model to output the numerical values relevant to the given ICO (here: \emph{remdesivir}, \emph{standard care}, \emph{all-cause mortality at up to day 28}). 

\begin{figure}[ht]
    \centering
    \includegraphics[scale=0.75]{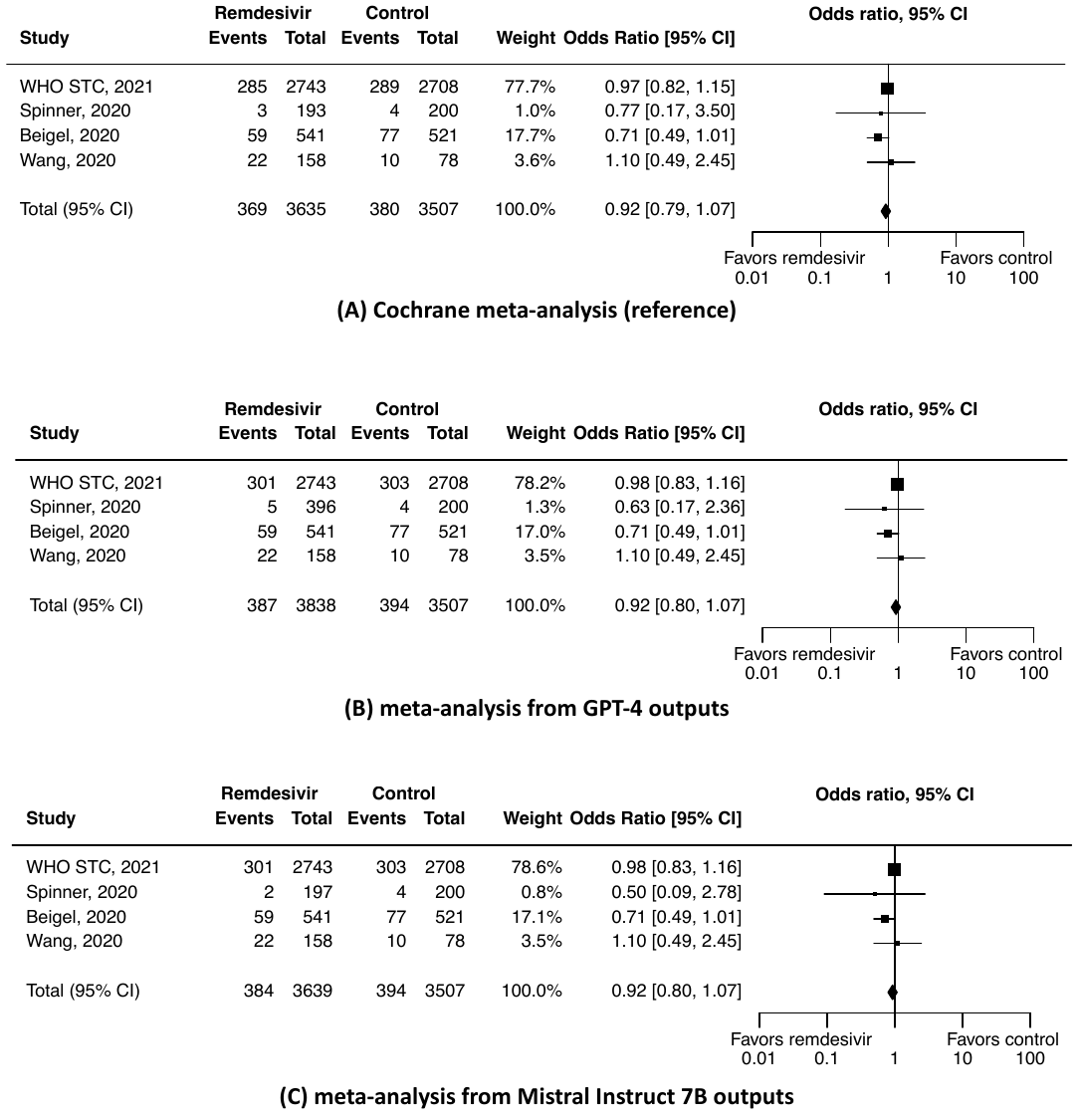}
    \caption{Forest plots showing fixed-effect meta-analyses from (A) Cochrane review (reference; performed with manually extracted data), (B) GPT-4 extracted data, (C) Mistral Instruct 7B extracted data. For this particular case, both LLMs permit highly accurate meta-analytic estimates, fully automatically.} 
    \label{fig:forest_plots}
\end{figure}

Both GPT-4 and Mistral correctly categorize the outcome type as binary. 
GPT-4 had an exact match accuracy of 0.500 and was able to match at least 2 values of the 2x2 outcome table correctly for all trials.  
Mistral performed comparably to GPT-4 in this case. 
The exact extracted data from the Cochrane reference and the two models are shown in \autoref{fig:forest_plots} as forest plots for comparison.\footnote{Forest plots show point estimates and associated confidence intervals corresponding to individual studies, as well as the aggregated (overall) estimate and interval.}
Both models made mistakes in extracting data from two studies \citep{who2021repurposed,spinner2020effect}. 
The errors for \citet{who2021repurposed} were related to the correct numbers not being readily available in the trial text but requiring extra calculations. 
For \citet{spinner2020effect}, GPT-4 hallucinated the events and the total numbers for the intervention, while Mistral extracted the wrong number due to choosing from the wrong timepoint for the outcome measure.
These errors were (very) minor when put into a fixed-effect meta-analysis; using outputs from both models resulted in the correct total log odds ratio and the 95\% CI being off by only 0.1 (\autoref{fig:forest_plots}).

\section{Discussion} 

In this work, we aim to assess whether modern LLMs are sufficiently capable of extracting data to produce accurate, fully automated meta-analyses. To this end, we comprehensively annotated a modest evaluation set of clinical trial reports with numerical findings attached to interventions, comparators, and outcomes. 
We then used this dataset to evaluate a wide range of LLMs in terms of their ability to infer structured numerical findings. 

Our results indicate that modern LLMs can perform this task with some accuracy, but difficulties remain. Massive general models like GPT-4 and GPT-3.5 perform fairly well, especially for tasks inferring outcome types and extracting binary outcomes. 
Smaller, open-source models also showed some capability but fared comparatively worse.
For some of the smaller models with minimal context windows, this may be because we had to pass chunks of the input trial reports through the network, increasing noise. 
Mistral 7B Instruct performed the best of the open-source models considered. Interestingly, despite being further trained on domain-specific text, PMC LLaMA and BioMistral performed considerably worse than other similar-sized general-purpose LLMs by producing the most number of ``unknowns'' and the lowest percentage of complete data. 
For extracting continuous outcomes, all the LLMs we evaluated performed below 50\% exact match accuracy.

We further demonstrated the use of LLMs in a case study in which we conducted a meta-analysis with automatically extracted results. 
This showed good results in almost replicating the exact total log odds ratio and confidence interval, suggesting a future in which meta-analysis is fully automatable. 
However, LLMs performed fairly well in this example probably because the constituent RCT reports have a clear structure in the outcome measure reports with limited ambiguity (such as different time points). 
Also, the example only contains 4 RCTs, which helps in reducing the accumulation of errors that can often happen with the inclusion of more data.

Our work shows that modern LLMs such as GPT-4 and Mistral are promising for extracting numerical findings from trial reports to produce automated meta-analyses. 
However, they fail to extract findings that require more specialized inference when multiple similar outcome measures are reported or the outcome measures are ambiguous. 
For now, these should remain an assistive technology, although it seems plausible that near-term improvements in LLMs may permit robust automatic meta-analysis.

\paragraph{Limitations}

There are several limitations to this work. Due to the time required to perform detailed annotations, the evaluation dataset we have introduced is a small sample. 
In addition, while the annotations were performed carefully by individuals familiar with meta-analysis, they did not have clinical background (that said, this work was done in consultation with a clinician to ensure accuracy).

We only investigated zero-shot applications of LLMs, mostly because of the context window that would be required to include examples of data extraction from full-texts. 
We did not attempt few-shot or fine-tuning.
Furthermore, we did only minimal and informal ``prompt engineering''; it is certainly possible that alternative instructions may yield better (or worse) performance.
A related limitation is that we tasked LLMs with inferring outcome types from only descriptions; providing additional context from articles may improve the performance on this first subtask.

% ACKNOWLEDGEMENTS ONLY GO IN THE CAMERA-READY, NOT THE SUBMISSION
\acks{This research was partially supported by National Science Foundation (NSF) grants RI-2211954 and IIS-1750978, and by the National Institutes of Health (NIH) under the National Library of Medicine (NLM) grant 2R01LM012086.}

%Do NOT change font size of references or modify the bibliography style
\bibliography{references}

\newpage
\appendix

\section*{Appendix A.}
\label{app:error_plots}

We present bar plots of the breakdown of the general error types we have identified from GPT-4 and Mistral Instruct 7B outputs. Our error analysis reveals that Mistral produces significantly more errors across all three tasks than GPT-4.

\begin{figure}[htbp]
    \centering
    \includegraphics[scale=0.6]{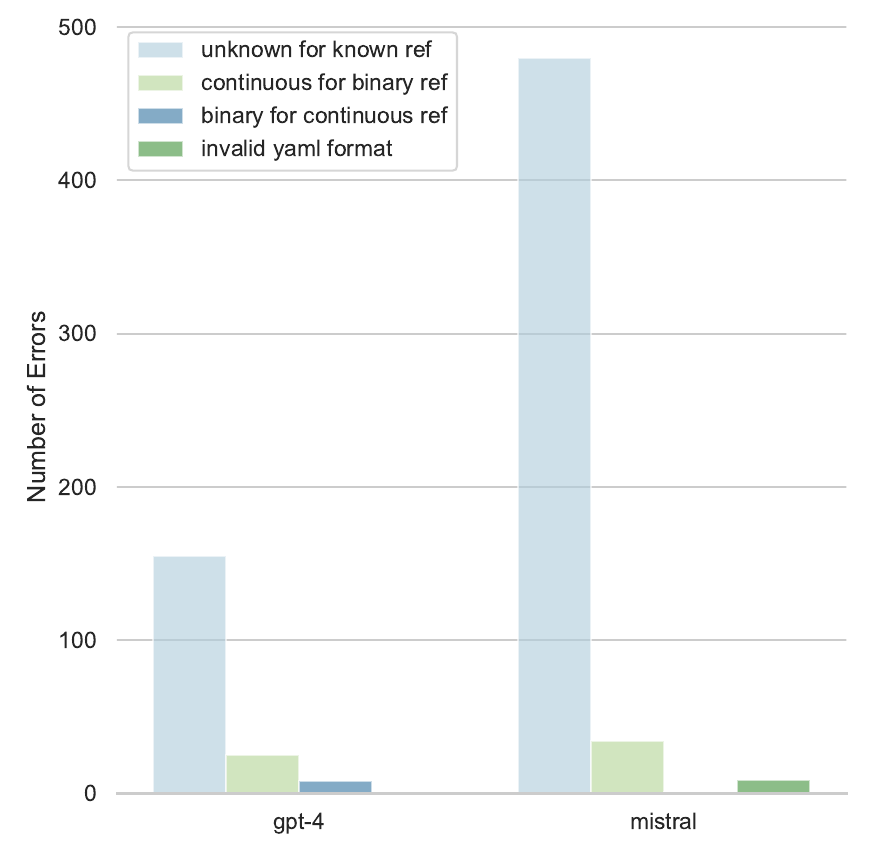}
    \caption{Bar plot showing the types of errors from GPT-4 and Mistral Instruct 7B for \emph{outcome type inference} task.} 
    \label{fig:outcome_type_errors_plot}
\end{figure}

\begin{figure}[htbp]
    \centering
    \includegraphics[scale=0.6]{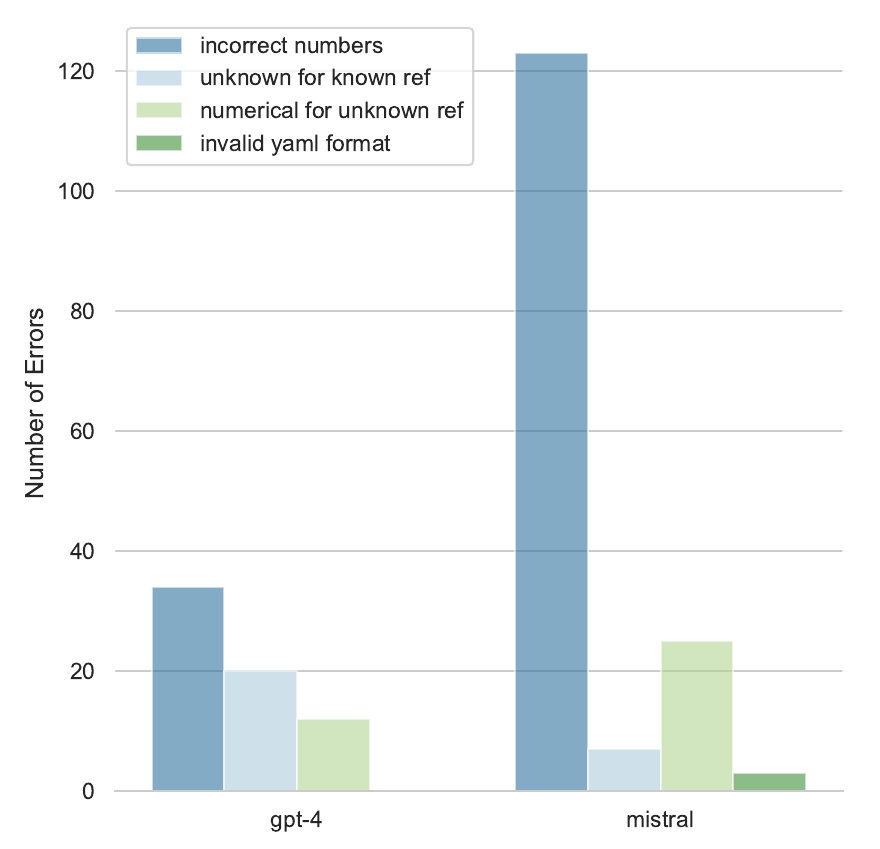}
    \caption{Bar plot showing the types of errors from GPT-4 and Mistral Instruct 7B for \emph{binary outcomes extraction} task.} 
    \label{fig:binary_outcomes_errors_plot}
\end{figure}

\begin{figure}[htbp]
    \centering
    \includegraphics[scale=0.6]{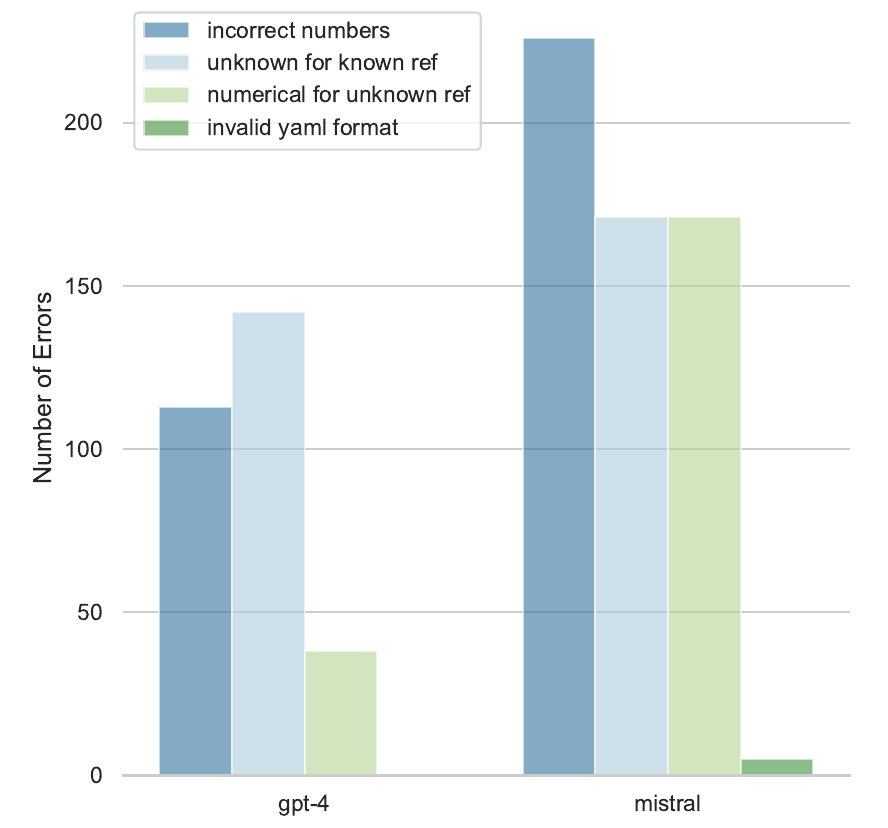}
    \caption{Bar plot showing the types of errors from GPT-4 and Mistral Instruct 7B for \emph{continuous outcomes extraction} task.} 
    \label{fig:continuous_outcomes_errors_plot}
\end{figure}

\clearpage

\section*{Appendix B.}
\label{app:samples}

Some samples of errors from GPT-4 and Mistral Instruct 7B are presented below. We sampled these as part of the qualitative error analysis from \hyperref[sec:results]{Section~\ref*{sec:results}}. 

\begin{table}[htbp]
\centering
\caption{Samples of errors from inferring outcome type with GPT-4 and Mistral Instruct 7B.}
\label{tab:outcome_type_errors_samples}
\begin{tabular}{llll}
\toprule
\textbf{Model} & \textbf{Outcome (Input)}        & \textbf{Reference} & \textbf{Output}\\ \midrule
\rowcolor[HTML]{EFEFEF} 
GPT-4          & communication ability          & continuous         & unknown         \\
GPT-4          & survival                       & continuous          & binary         \\
\rowcolor[HTML]{EFEFEF} 
GPT-4          & differences in disc herniation & binary      & continuous             \\
Mistral Instruct & glycemic level                 & continuous         & unknown         \\
\rowcolor[HTML]{EFEFEF} 
Mistral Instruct & completed treatment            & continuous          & binary         \\
Mistral Instruct & level of postoperative pain    & binary      & continuous             \\ \bottomrule
\end{tabular}
\end{table}

\begin{figure}[htbp]
    \centering
    \includegraphics[scale=0.56]{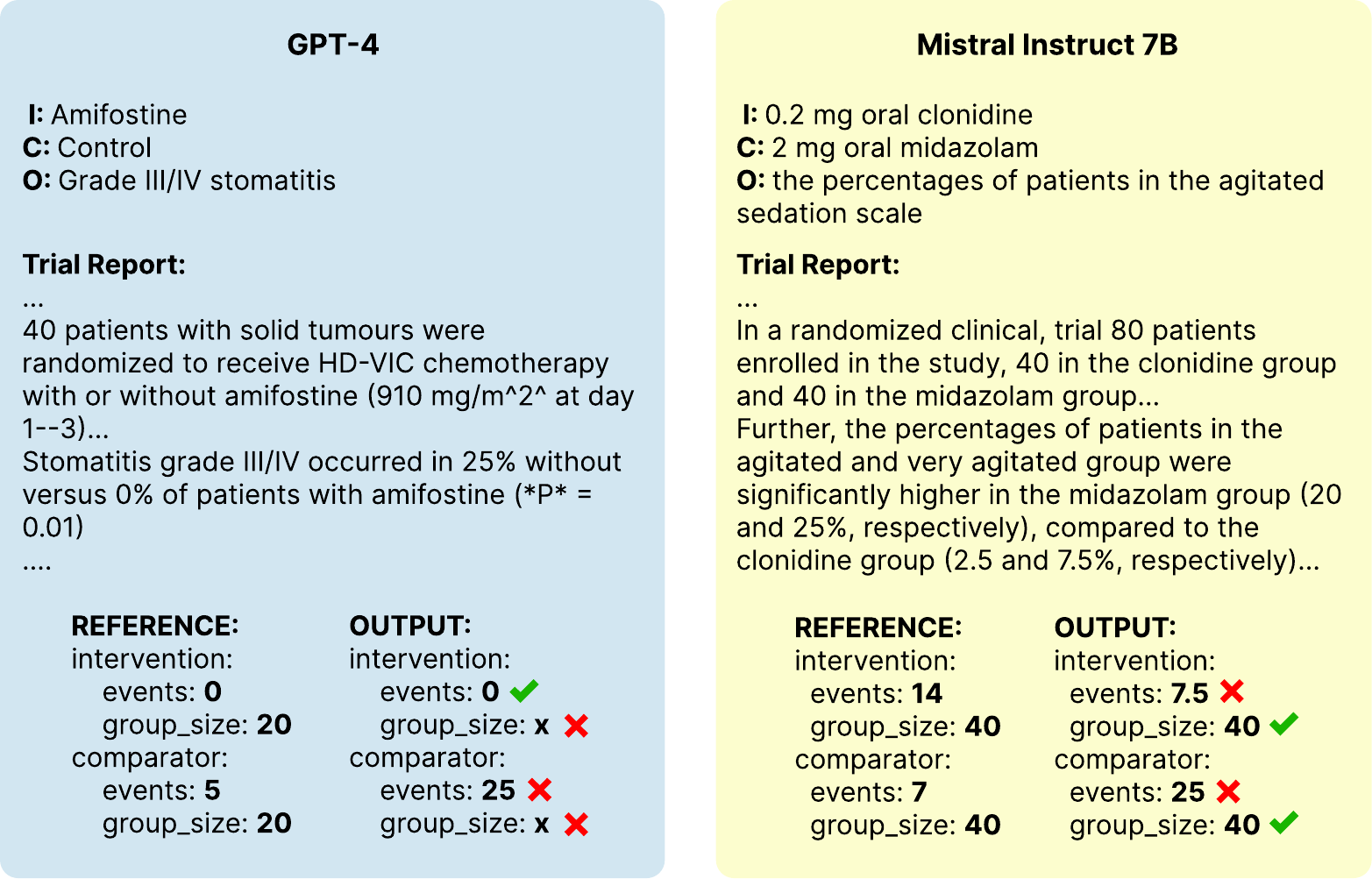}
    \caption{Samples of errors from extracting binary outcomes with GPT-4 and Mistral Instruct 7B. The GPT-4 error is from failing to infer that the group sizes should be half the total number of participants. For Mistral, the model received a binary outcome with the word ``percentages.'' Mistral was faithful to the given outcome by extracting the percent numbers found in the text but failed to output participant numbers which are more often used for binary outcomes.} 
    \label{fig:binary_outcomes_errors_samples}
\end{figure}

\begin{figure}[htbp]
    \centering
    \includegraphics[scale=0.56]{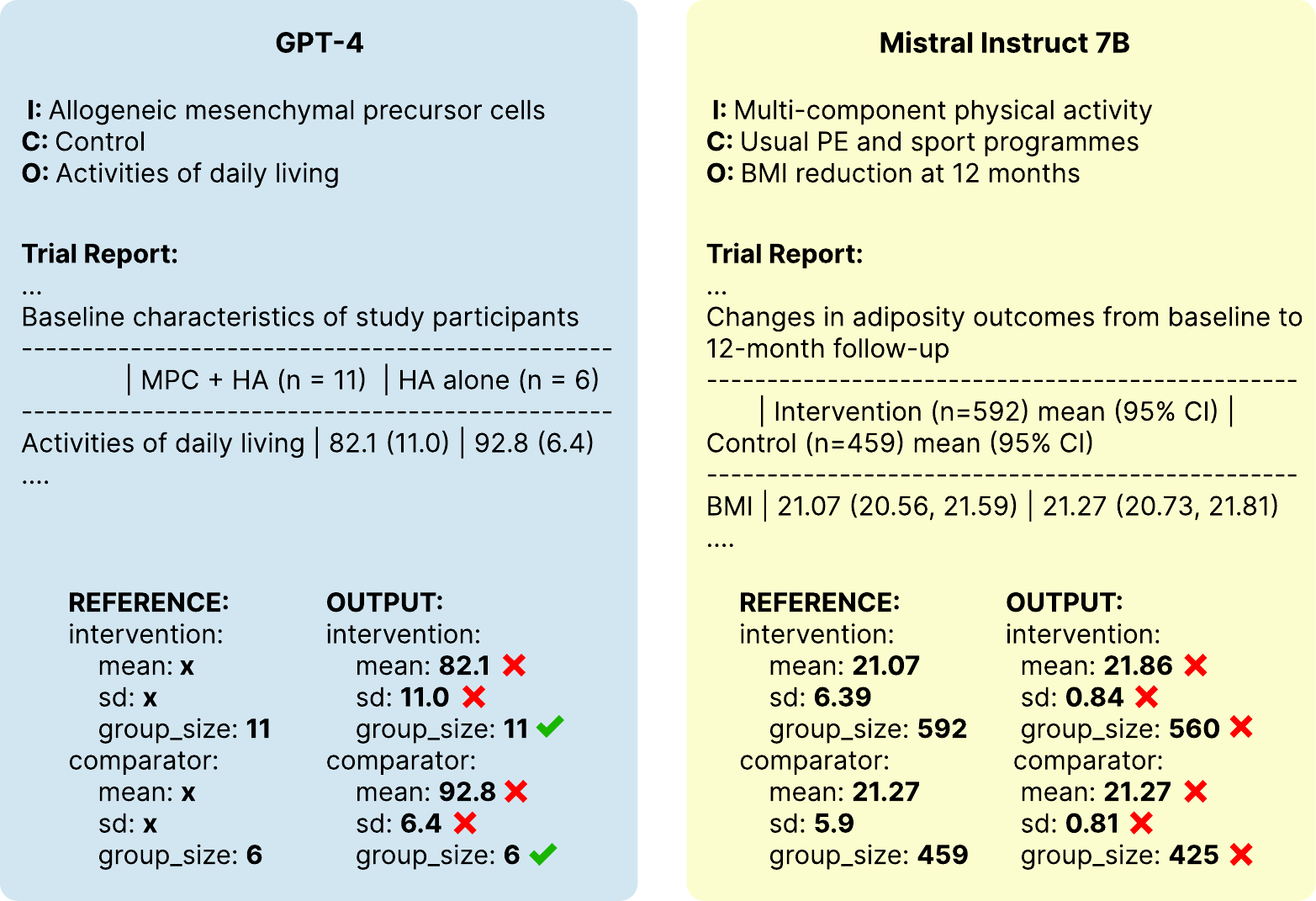}
    \caption{Samples of errors from extracting continuous outcomes with GPT-4 and Mistral Instruct 7B. The error from GPT-4 is due to the model extracting values from the baseline and not the post-intervention which is the outcome measure of interest. The error from Mistral is related to the model extracting the means and standard deviations from the wrong locations of the text.} 
    \label{fig:continuous_outcomes_errors_samples}
\end{figure}

\end{document}